\documentclass{article}

\usepackage[final, nonatbib]{crl_neurips_2023}

\usepackage[utf8]{inputenc} % allow utf-8 input
\usepackage[T1]{fontenc}    % use 8-bit T1 fonts
\usepackage{hyperref}       % hyperlinks
\usepackage{url}            % simple URL typesetting
\usepackage{booktabs}       % professional-quality tables
\usepackage{amsfonts}       % blackboard math symbols
\usepackage{nicefrac}       % compact symbols for 1/2, etc.
\usepackage{microtype}      % microtypography
\usepackage{xcolor}         % colors
\usepackage{amsmath}
\usepackage{amsfonts}
\usepackage{multirow}
\usepackage{graphicx}
\usepackage[toc,page]{appendix}
\usepackage{subcaption}
\usepackage{floatrow}
\usepackage{tablefootnote}
\title{SCADI: Self-supervised Causal Disentanglement \\ in Latent Variable Models}

\author{%
Heejeong Nam \thanks{\url{https://hazel-heejeong-nam.github.io/}}\\
Department of Electrical and Electronic Engineering\\
Yonsei University\\
Seoul, South Korea \\
\texttt{hatbi2000@yonsei.ac.kr} \\
}
\begin{document}

\maketitle

\begin{abstract} %길이 완료, 문법 완료
Causal disentanglement has great potential for capturing complex situations. However, there is a lack of practical and efficient approaches. It is already known that most unsupervised disentangling methods are unable to produce identifiable results without additional information, often leading to randomly disentangled output. Therefore, most existing models for disentangling are weakly supervised, providing information about intrinsic factors, which incurs excessive costs. Therefore, we propose a novel model, \textit{SCADI(SElf-supervised CAusal DIsentanglement)}, that enables the model to discover semantic factors and learn their causal relationships without any supervision. This model combines a masked structural causal model (SCM) with a pseudo-label generator for causal disentanglement, aiming to provide a new direction for self-supervised causal disentanglement models.
\end{abstract}
\section{Introduction} %문법 완료, 한줄 길어진듯, especially 이상함함
\begin{figure}[tb!]
\centering
\includegraphics[width=8cm]{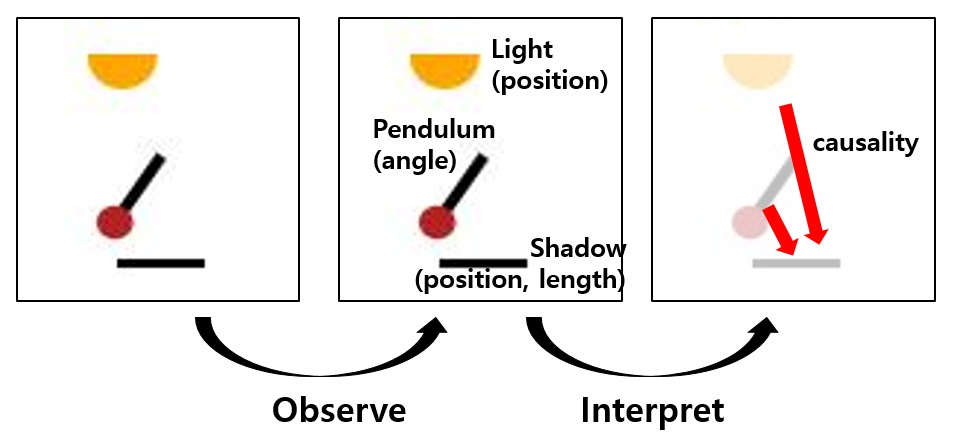}
\caption{Process of understanding image : Observation and Interpretation.}
\label{learning}
\end{figure}

Imitating humans has been the ultimate goal of machine learning, and now machine learning is capable of performing various tasks. However, due to the inherent drawbacks of black box models, there are still limitations in understanding and learning complex relationships. To imitate image understanding of human, we consider a two-step process, as depicted in Fig. \ref{learning}. The first step is observation, learning about the various elements presented in the data, such as light sources, a swinging pendulum, and shadows (see Fig. \ref{learning}). The second step is interpretation, which involves understanding the relationships among the elements identified during the observation stage. This paper aims to propose a methodology that can perform both observation and interpretation without any supervision. Our approach, \textit{SCADI (Self-supervised Causal Disentanglement)}, is based on disentangled representation learning, and brings us closer to the process of human thinking. Disentangling aims to understand the factors of variation in the data, and can compress complex data in a concise and information-rich manner, making it beneficial for downstream tasks \cite{bengio2013, lecun2015, good2009, locatello2019fairness, Suter2018InterventionalRO, higgins2018definition, bengio2007}. The early-stage disentangling was done through unsupervised learning using latent variable models with an independence assumption on factors, such as variational autoencoders (VAEs)\cite{betavae, factorvae, kumar2018variational, betatcvae}. Among them, $\beta$-VAE \cite{betavae}\cite{betavae_ex} serves as our baseline model for \textit{observer}. $\beta$-VAE adjusts the balance between the reconstruction loss and the Kullback-Leibler (KL) divergence in the objective function. Minimizing the KL divergence loss, weighted by $\beta$, enforces independence among factors by encouraging the latent variables to align with the prior distribution. However, Locatello et al.\cite{localimpossible} demonstrated that unsupervised disentangling lacks identifiability \cite{identi}, making it difficult to get consistent results. As a result, many weakly supervised disentangling models emerged \cite{laddervae}\cite{denton2017unsupervised}\cite{li}\cite{kulkarni2015deep} \cite{bouchacourt2017multilevel}, which incorporate additional information or inductive biases . Nevertheless, the cost of the labels is excessive. Therefore, various semi-supervised disentangling approaches have been proposed\cite{reed} \cite{cheung2015discovering} \cite{mathieu2016disentangling} \cite{localfewlabel}, with many still maintaining the independence assumption. However, feature independence is often unrealistic (Fig. \ref{learning}). Consequently, models such as CausalGAN\cite{kocaoglu2017causalgan}, CausalVAE\cite{causalvae}, and DEAR\cite{dear} have emerged, which have discarded the independence assumption, but they still provide additional information in the form of labels\cite{causalvae} or by incorporating prior causal graphs \cite{kocaoglu2017causalgan}\cite{dear}, or using weakly paired datasets\cite{weaklysth}. DEAR\cite{dear} aims to utilize a practical amount of labeled data, but it requires prior knowledge about causal graph.

%문법 됨 길이 됨
To significantly alleviate the burdens of supervision, we propose the first self-supervised approach that prevents random disentanglement. In our best knowledge, SCADI(Self-supervised CAusal Disentanglement), is the first attempt at achieving causal structured disentanglement without any additional information or inductive biases. SCADI has two main components: 1. An \textit{observer}, which performs dimension-wise unsupervised disentangling through a latent variable model, generating pseudo-labels. 2. An \textit{interpreter}, a module for vector-wise weakly supervised causal disentanglement. It relies on the structured causal model (SCM) \cite{scm_itself,masked_scm}. SCM has played a significant role in incorporating causality into latent variable models\cite{dear}\cite{he2018variational}\cite{moraffah2020causal}\cite{kocaoglu2017causalgan}, including in CausalVAE\cite{causalvae}, which serves as the baseline for \textit{interpreter}. In SCADI, the \textit{interpreter} is supervised by the labels generated by the \textit{observer}, while the \textit{observer} receives additional regularization from the \textit{interpreter}, which forces the adjacency matrix in masked SCM to be a directed acyclic graph (DAG). Our detailed explanations and experiments address how SCADI performs causal disentanglement. Our code is available at \url{https://github.com/Hazel-Heejeong-Nam/Self-supervised-causal-disentanglement}.

\section{Method}
%문법 됨 길이 됨
Starting with \textit{observer}, we adopt a latent variable model for unsupervised disentangling, $\beta$-VAE\cite{betavae}, to utilize the latent space as pseudo-labels. While $\beta$-VAE, does not capture causality directly, its KL divergence forces it to focus on the most distinguishable features in data. However, as highlighted by Locatello et al.\cite{localimpossible}, unsupervised disentangling models not only cannot always produce well-disentangled results, but also lack the ability to disentangle correlated factors. This prompted us to consider providing additional regularization to improve the quality of the pseudo-labels.

\begin{figure*}[tb!]
\centering{\includegraphics[width=\textwidth]{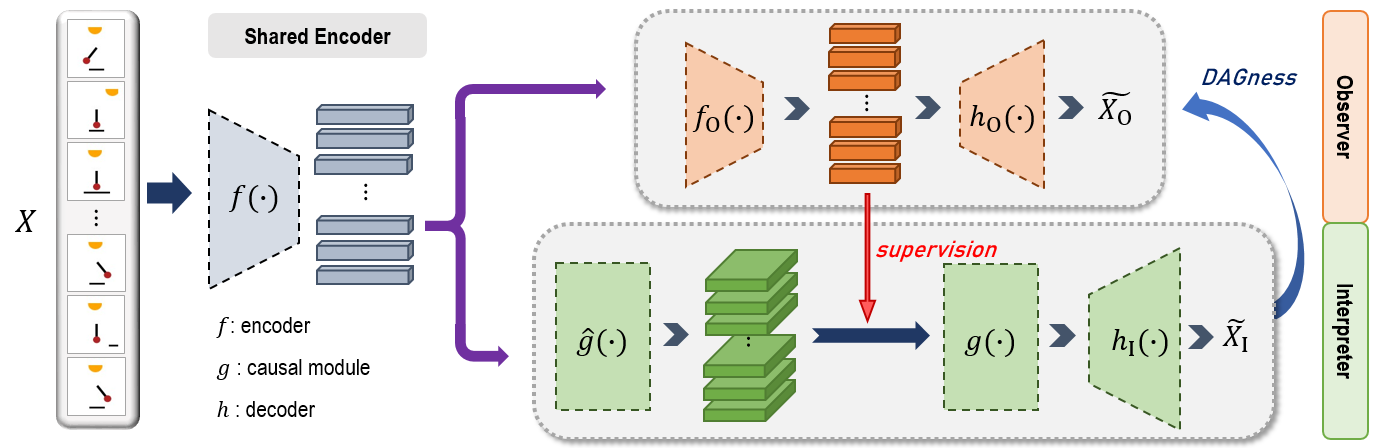}}
\caption{Overview of \textbf{self-supervised structured causal disentanglment}.}
\label{fig:mainmodel}
\end{figure*}

%여기부터 엔터까지 문법 됨 길이 됨
\textbf{Definition 1} (Symbols of our model). \textit{We will begin by defining our notations. $\hat{g}(\cdot)$ and $g(\cdot)$ will be also explained further in \ref{section:interpreter}}
\begin{itemize}
    \item[1. ]\textit{\textbf{Dataset} $X$ consists of $n$ images. i.e. $X:=\{x_1,...,x_n\}$, where $x_i \in \mathbb{R}^{W\times H\times C}$. }
    
    \item[2. ]\textit{\textbf{Adjacency Matrix} $\mathbb{A}:=\{A_1,...,A_n\}$, where  $A_i:=[A_{i}^1|...|A_{i}^c]$ and $A_{i}^{jk}:=\{A_i^{jk}\}_{k=1}^c$. $c$ is the number of factors of interest in data.}

    \item[3. ]\textit{\textbf{Exogenous latent variables} are $E:=\{\epsilon_1,...,\epsilon_n\}$, where $\epsilon_i:=f(x_i)$. $f(\cdot)$ is the first encoder in Fig.\ref{fig:mainmodel}. $\epsilon_i \in \mathbb{R}^{\alpha \cdot c}$ where $\alpha$ is an arbitrary positive number. \textbf{Endogenous latent variables} are $z_i=\hat{g}(A_i,\epsilon_i)$ , $z_i\in \mathbb{R}^{c\times \alpha}$and $Z:=\{z_1,...z_n\}$, where $z_i^{jk}$ is value of $j^{th}$ row and $k^{th}$ column in $z_i$.}
    
    \item[4. ]\textit{\textbf{Observed Labels} are $U:=\{u_1,...,u_n\}$, where $u_i:=f_O(\epsilon_i)$, are defined as observed labels from Observer. $f_O$ is an additional \textit{observation} encoder followed by the shared encoder, and $u_i:=\{u_{i}^k\}_{k=1}^{c}$.}
    
    \item[5. ]\textit{\textbf{Decoders} : Observation decoder and interpretation decoder are denoted as $h_O(\cdot)$ and $h_I(\cdot)$ respectively. We define reconstructed data as $\tilde{x_i}_O:=h_O(u_i)$ and $\tilde{x_i}_I:=h_I(g(A_i,z_i)$}
\end{itemize}

\begin{figure}[htb!]
\centering{\includegraphics[width=9cm]{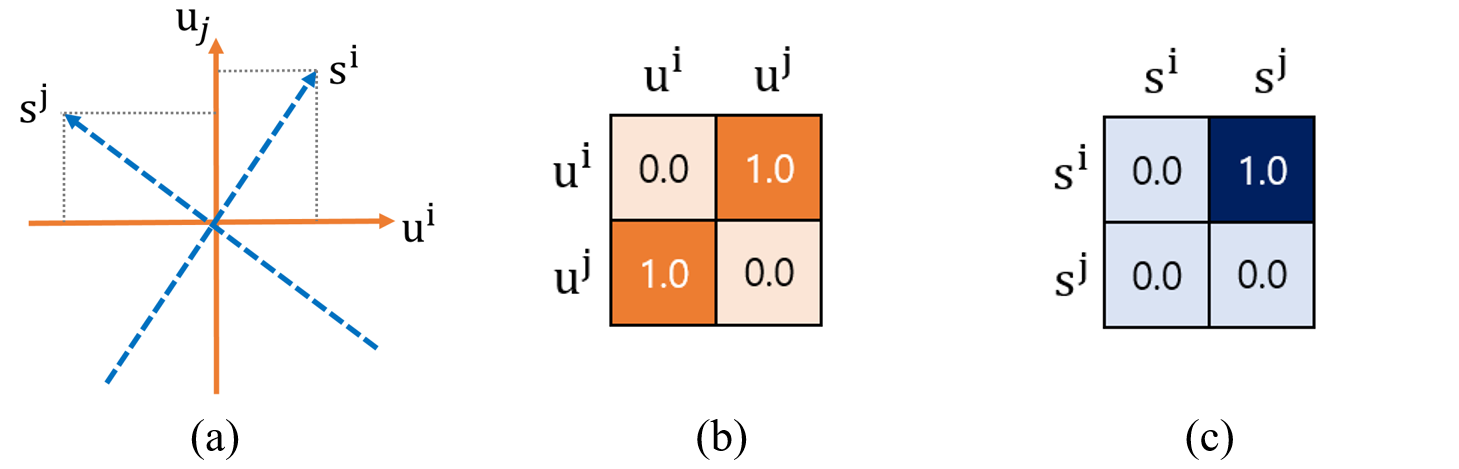}}
\caption{(a) Observed factors and true underlying factors are entangled. (b) Adjacency matrix of observation. (c) Adjacency matrix of true factors.}
\label{fig:dagimportant}
\end{figure}

\textbf{Example 1} . \textit{We additionally define true underlying factor set $S$ with $c$ factors in it, i.e. $S:=\{s^1,...,s^c\}$. Any pair of $(s^i, s^j)$ can be either independent or causally related. Fig. \ref{fig:dagimportant} (a) shows an entangled case of the \textbf{observer}'s disentangling process. Both observed factors $u^i$ and $u^j$ are having combined effects of true factors $s^i$ and $s^j$. Here, we assume $i^{th}$ factor is a cause of $j^{th}$ factor, i.e. $i^{th}$ factor and $j^{th}$ factor are in \textbf{parent-child} relationship. Successful disentanglement would make the distributions of true underlying factor $S$ and observed factor $U$ aligned.}

We incorporate the concept of \textbf{\textit{DAGness}}\cite{causalvae,zheng2018dags,yu2019daggnn} in the adjacency matrix for SCM, denoted as $H(A)$ in \eqref{eq:dagness}. After the labels were passed to the \textit{interpreter}, SCADI performs causal disentangling within SCM and learns $A$. In \textbf{Example 1}, $A$ of true underlying factors should be like Fig.\ref{fig:dagimportant}(c). However, if the generated labels are entangled, as shown in Fig.\ref{fig:dagimportant}(a), $A$ would exhibit bidirectional relationships, as in Fig.\ref{fig:dagimportant}(b). Therefore DAGness of $A$ can assess the disentanglement in observer, and in the same context, minimize DAGness would assist disentangling factors by helping to suppress bidirectional relationships and anchor the factors in place. Although this constraint does not yet achieve complete mathematical identifiability\cite{identi}, we insist that DAGness prevents randomly disentangled results in the \textit{observer}. The proved identifiability of the CausalVAE\cite{causalvae}, corresponding to our interpreter, implies that if only the generated labels are well-disentangled, the identifiability of SCADI will also be satisfied.

\begin{equation}
\label{eq:dagness}
H(A) \equiv tr((I + A\circ A)^c)-c
\end{equation}

\begin{figure}[tb!]
\centering{\includegraphics[width=13cm]{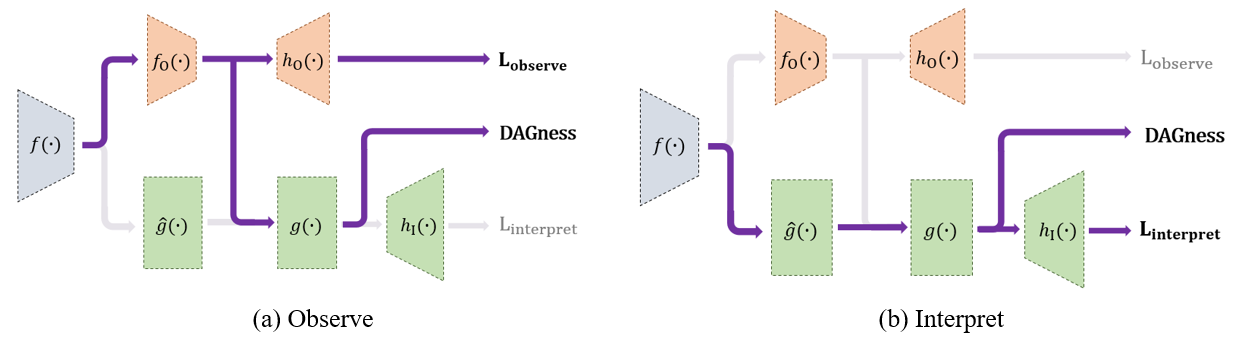}}
\caption{Gradient flow while training \textit{observer} and \textit{interpreter}.}
\label{fig:grad}
\end{figure}

%여기부터 엔터까지 길이 됨 문법 됨
\paragraph{Observer and Interpreter}
\label{section:interpreter}
The task of the \textit{observer} is to provide a scalar label for each factor. \textit{Observer} has two key differences from $\beta$-VAE\cite{betavae}. Firstly, it undergoes a two-step encoding process. In the first encoding step, $f(\cdot)$ shares its weights and the latent space with the \textit{interpreter}. The exogenous latent vector $E$ from $f(\cdot)$ does not inherently encode relationships among factors. In the second step, the output vector $U$ with a fixed length $c$ is obtained through $f_O(\cdot)$. $U$ is not only fed to the decoder $h_O(\cdot)$ but also serves as the label passed to the \textit{interpreter}. Finally, by adding the evidence lower bound (ELBO) loss, the objective function of the \textit{observer} can be written as \eqref{eq:observer}.
\begin{align}
L_{obs} &= -\mathbb{E}_{q_\phi (u|x, \epsilon)}[ \log p_\theta (x,\epsilon|u)] + w_O^d H(A) + \beta D_{KL}(q_\phi (u|x,\epsilon)||p_\theta(u)) \nonumber \\ \label{eq:observer}
&= -\text{ELBO} + w_O^dH(A)
\end{align}
The \textit{interpreter} is similar to CausalVAE\cite{causalvae}, which adopts SCM. The exogenous latent variable $E$ is mapped to the endogenous latent variable $Z$ through causal inference, using the adjacency matrix $A$. Here, a linear SCM\cite{scm_itself} equation, as shown in \eqref{inf}, is employed. We defined \eqref{inf} as $\hat{g}$. Subsequently, SCADI performs causal disentanglement within the masked SCM\cite{masked_scm}. By masking out non-parental elements of $Z$ using $A$ in each semantic vector, the model is able to learn the effects of the individual factors while maintaining their connections with their parental semantics. This can be written as \eqref{eq:mask}, where $j$ represents each concept and $a$ is a non-linear function for stability. We defined \eqref{eq:mask} as $g$, where $\eta_i$ is the parameter set of $a$.
\begin{equation}
\label{inf}
z = A^Tz + \epsilon = (I-A^T)^{-1} \epsilon, \quad \epsilon \sim N(0,I)
\end{equation}
\begin{equation}
\label{eq:mask}
z^j = a^j(A^j \odot z; \eta^j)+\epsilon^j
\end{equation}
\begin{equation}
\label{eq:label}
u^j = a^j(A^j \odot u; \eta^j)
\end{equation}
%길이됨 문법됨됨
In Similar way, the labels generated from the \textit{observer} are fed to the SCM layer as \eqref{eq:label}. The mask loss, denoted as $l_m$, compares $z$ before and after applying \eqref{eq:mask}, and then incorporated into the objective function of the \textit{interpreter}. Similarly, the label loss, $l_u$, can be obtained by comparing $u$ before and after the mask layer. Adding evidence lower bound (ELBO) loss, the \textit{interpreter} loss can be described as \eqref{eq:interpreter}.  We follow Yang et al.\cite{causalvae} for further details. To summarize, the \textit{observer} receives DAGness from the \textit{interpreter}, while the \textit{interpreter} obtains labels from the \textit{observer}, establishing a mutually beneficial relationship. During training, in order for all parameters to be updated at least once, two forward passes are needed. The first pass aims to minimize the objective function for the \textit{observer}\eqref{eq:observer}, as illustrated by the gradient flow depicted in Fig.\ref{fig:grad}(a). The second forward pass aims to minimize the objective function for the \textit{interpreter}\eqref{eq:interpreter}, and we detach the gradient of the label as in Fig.\ref{fig:grad}(b). During inference, only a single forward pass is required.
\begin{align}
L_{int} &= -\mathbb{E}_{\epsilon, z}[\log p_\theta (x|z, \epsilon, u)] + D_{KL}(q_\phi(\epsilon, z|x,u)||p_\theta(\epsilon,z|u)) + w_I^d H(A) + w_I^u l_u + w_I^m l_m  \nonumber \\ \label{eq:interpreter}
&= -\text{ELBO} + w_I^d H(A)+ w_I^u l_u + w_I^m l_m
\end{align}
%길이됨 문법됨됨
\section{Experiments}
\textbf{Synthetic pendulum dataset} \quad We utilized the \textbf{synthetic pendulum dataset} by Yang et al. \cite{causalvae}. Each image consists of a light source, a pendulum, and a shadow with varying lengths and locations determined by the position of the light source and pendulum. The factor of variants are as follows\cite{causalvae}\cite{dear} : \textit{1) pendulum angle, 2) light position, 3) shadow length, and 4) shadow position.} With the official split of the train and test set\cite{causalvae}, we obtained 5482 training images and 1826 test images. Fig.\ref{fig:doset}(b) shows the true causal graph of the described factors.

\textbf{Baselines} We compared our model with 3 differenct architectures: CausalVAE\cite{causalvae}, unsup-CausalVAE\cite{causalvae}, and nd-SCADI. Unsup-CausalVAE eliminates supervision from the masked SCM as Yang et al.\cite{causalvae} did, which is equivalent to using only \textit{interpreter} in SCADI. Nd-SCADI is a modified model from SCADI, which abandoned extra DAGness regularization to the \textit{observer}. Table \ref{tab:ox} shows differences among SCADI and its baselines. We also provide brief structures of our baselines in Appendix \ref{appendix:exp}.

\begin{table}[ht!]
\ttabbox[]
  {\centering
   \tabcolsep=0.1cm
    \renewcommand{\arraystretch}{1.3}
    \small
    \begin{tabular}{cccc}
    \hline
                    & \textbf{Supervision} & \textbf{Observer} & \textbf{Interpreter} \\ \hline \hline
    CausalVAE\cite{causalvae}       & $\circ$           & $\times$        & $\circ$           \\ \hline
    unsup-CausalVAE\cite{causalvae} & $\times$           & $\times$        & $\circ$           \\ \hline
    nd-SCADI        & $\times$           & $\circ$        & $\circ$           \\ \hline
    SCADI           & $\times$           & $\circ$        & $\circ$           \\ \hline
    \end{tabular}}%
    {\caption{Baselines and architecture summary}\label{tab:ox}}%
\end{table}%

\subsection{Setup and evaluation}

\begin{figure}[tb!]
\centering{\includegraphics[width=13cm]{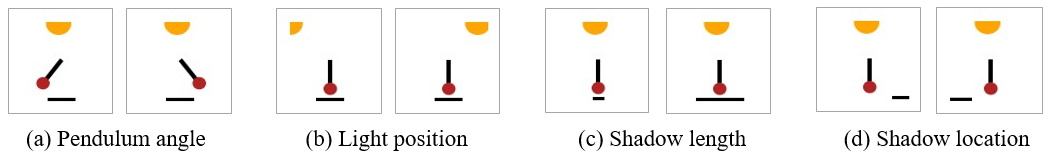}}
\caption{Label-finding set. (a),(b),(c),(d) are showing paired data for corresponding factors.}
\label{ffset}
\end{figure}

For the \textit{observer} in SCADI, we assess how well the model separates the factors. In existing weakly supervised methods, providing supervision makes it easier to determine which dimension of the latent space encodes a specific underlying factor. However, since unsupervised disentangling does not have a predefined order in the latent vector, it becomes essential to reveal the order of the factors for further \textit{interpreter} analysis. To do so, we categorized a small subset of counterfactual images, twice the number of factors, as the label-finding set depicted in Figure \ref{ffset}. We then used these images to examine which factors are encoded in each dimensions of the latent vector produced by \textit{observer}.

%길이됨 문법됨됨
\textbf{Definition 2} (Label-Finding process). \quad \textit{As shown in Fig. \ref{ffset}, the label-finding set is divided into subgroups. Each subgroup consists of a pair of counterfactual images, $(F_a^i,F_b^i)$, which differ only in the state of the $i^{th}$ semantic, while the other semantics remain the same. By calculating the absolute element-wise difference between the latent variables generated from $F_a^i$ and $F_b^i$, we can quantify the variation in each dimension when the $i^{th}$ semantic changes. We consider the index of the latent variable with the largest difference as $label_i$, since the $i^{th}$ value in the latent vector strongly encodes the $i^{th}$ semantic. \eqref{ffeq} summarizes the label-finding process, where \textbf{obs} denotes the observation process of getting the labels.}

%길이됨 문법됨됨
\noindent
\textbf{Definition 3} (Label-Quality score). \quad \textit{We defined the LQ(Label-Quality) score measured based on \textbf{Label-Finding process}. We calculate cross-entropy loss between label and $|obs(F_{a}^i)- obs(F_{b}^i)|$. Eq. \eqref{eq:labelscore} shows how LQ score is calculated. We consider a lower LQ score to indicate better performance.}
\begin{equation}
\label{ffeq}
label_i = \text{max} (|obs(F_{a}^i)- obs(F_{b}^i)|)
\end{equation}
\begin{equation}
\label{eq:labelscore}
lq_i = \text{Cross Entropy} (|obs(F_{a}^i)- obs(F_{b}^i)|, label_i)
\end{equation}
We prioritize non-overlapping labels. Even if the LQ score is better, overlapped labels indicate suboptimal performance. Our evaluation enables an examination of how semantics are represented and how strongly they are disentangled. For the evaluation of the \textit{interpreter}, we followed CausalVAE\cite{causalvae}. Quantitatively, we examine the DAGness where a smaller DAGness indicates a less entangled result. Qualitatively, we first directly compared the obtained causal graph to the ground truth. While each value in $A$ ranges from 0 to 1, we rounded up to determine causality. Secondly, since most of the casual disentangling models are generative latent variable models\cite{causalvae, kocaoglu2017causalgan, dear,moraffah2020causal}, learned causality can be visualized through do-operations\cite{dear}\cite{causalvae}, which intervene on latent variables to generate counterfactual data \cite{besserve2019counterfactuals}. Ideally, if the parent element is changed, the corresponding child also should be changed accordingly, while the parent element should remain unaffected even though the child element has been changed. See Appendix \ref{appendix:imp} for our implementation details.
%길이됨 문법됨됨
\subsection{Experiment results}
\begin{figure}\BottomFloatBoxes
\floatsetup{captionskip=4pt}
\begin{floatrow}
\ttabbox[]
  {\centering
   \tabcolsep=0.05cm
    \renewcommand{\arraystretch}{1.4}
    \small
    \begin{tabular}{ccccc}

    \hline
     \textbf{Intervene}               & $u[0]$   & $u[1]$   & $u[2]$   & $u[3]$   \\ \hline \hline
    shadow length          & \textbf{1.6296} & 0.9784 & 0.1379 & 0.5790 \\ \hline
    light           & \textbf{7.0160} & \textit{6.8370} & 3.3080 & 0.3135 \\ \hline
    pendulum        & 2.6932 & 0.9545 & \textbf{3.6941} & 2.8357 \\ \hline
    shadow location & 0.7641 & 0.1960 & 1.8232 & \textbf{3.3046} \\ \hline 
    \end{tabular}}%
    {\caption{Label finding process}
    \label{tab:ours}}%

    \killfloatstyle\ffigbox[]{
        \includegraphics[width=\linewidth]{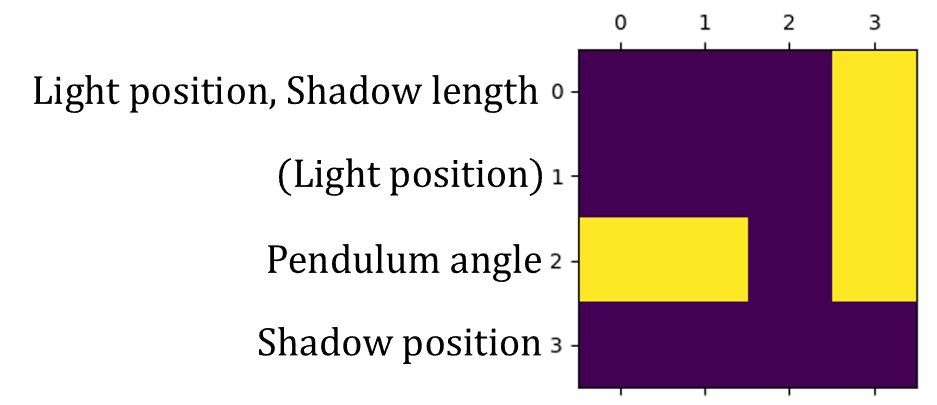}
    }
    {
        \caption{Adjacency matrix of SCADI}
        \label{fig:oursA}
    }
\end{floatrow}
\end{figure}
\begin{figure}[tb!]
\centering{\includegraphics[width=11cm]{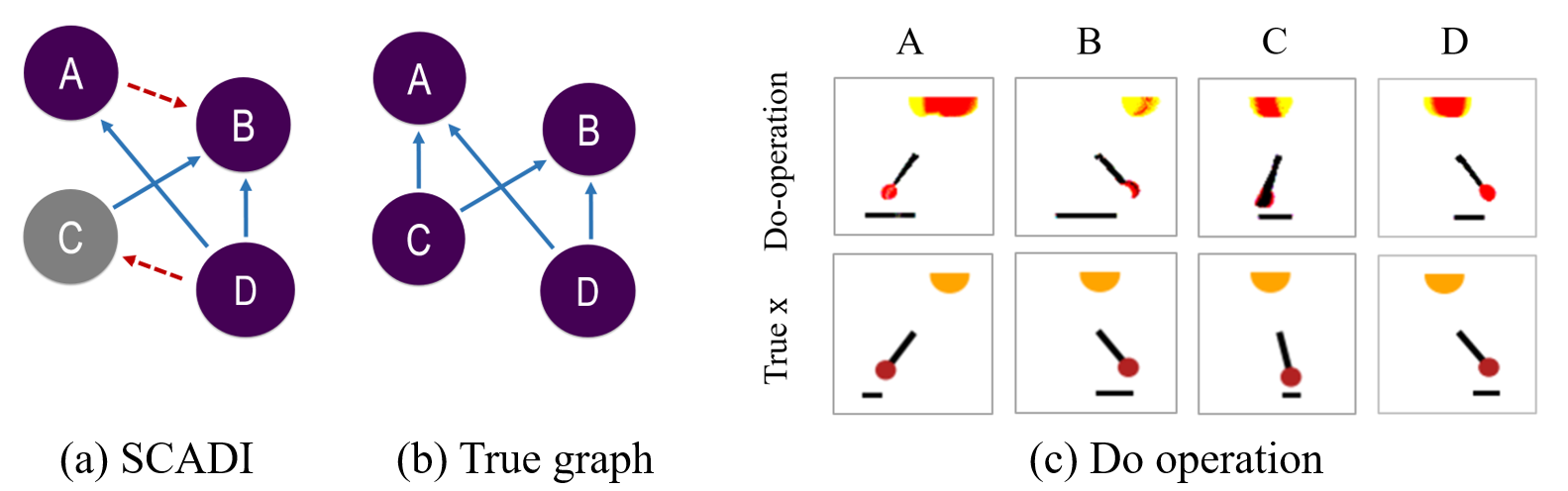}}
\caption{For simplicity, following symbols are used: \textbf{A} for the shadow length, \textbf{B} for the light position, \textbf{C} for the pendulum angle, and \textbf{D} for the shadow location. (a) Graph based on Fig. \ref{fig:oursA}. Red arrows indicate a incorrect causal relationship. \textbf{C} is ambiguous in the result of SCADI. (b) True causal graph. (c) Result of do-operation. To lessen ambiguity, the \textbf{A} is considered to be in $u[0]$ and \textbf{C} is in $u[1]$.}
\label{fig:doset}
\end{figure}

%
\paragraph{Observation and interpretation} \label{sec:oandi} Table \ref{tab:ours} shows $|obs(F_{a}^i)- obs(F_{b}^i)|$ of SCADI while intervening each factor. \textit{Pendulum angle} and \textit{shadow location} are fully disentangled in $u[2]$ and $u[3]$ respectively, having the same largest value both in row and column. However, \textit{shadow length} and \textit{light position} have the largest value in $u[0]$ in their row, thus seem to be entangled. Even though they are not fully disentangled, column $u[1]$ strongly encodes light position. Consequently, we labeled each latent dimension as Fig.\ref{fig:oursA}. Then we compare our obtained result (Fig. \ref{fig:doset} (a)) to the true causal graph (Fig. \ref{fig:doset} (b)). We found that SCADI can capture causal relationship to some extent in a cost-effective manner. Fig. \ref{fig:doset} (c) demonstrates that SCADI generates strong labels which are able to reconstruct counterfactual images by do-operation. In detail, \textbf{A} and \textbf{C} intervene light position and pendulum angle respectively, which are both parental elements of the shadow position and length. Thus, the position and length of the shadow changed as a result of do-operation. However, in \textbf{B} and \textbf{D}, the length and center location of the shadow are changed respectively, which are not parental elements for none of the others. Thus, even when shadow labels are intervened, the other factors remain the same. 

%길이됨 문법됨됨

\begin{figure}[b!]
\centering{\includegraphics[width=0.8\textwidth]{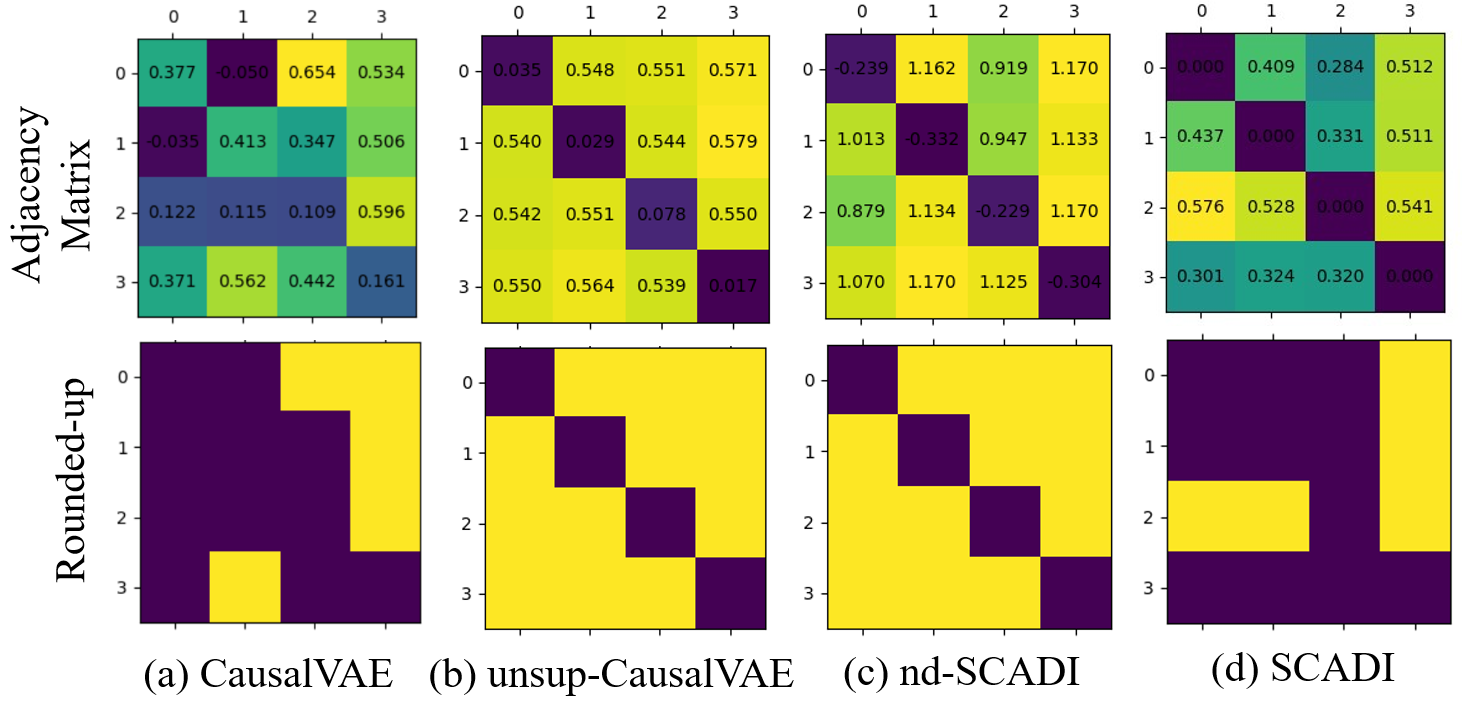}}
\caption{Comparison of the adjacency matrices of various unsupervised models.}
\label{fig:exp2}
\end{figure}

\paragraph{Comparative Evaluation}\label{sec:unsup} Fig.\ref{fig:exp2} illustrates the final adjacency matrices of four models (Table \ref{tab:ox}), including SCADI. It is evident that the unsup-CausalVAE and nd-SCADI exhibit subpar performance, as evidenced by the full entanglement within adjacency matrix. In contrast, CausalVAE with weak supervision and SCADI with self-generated labels show adjacency matrices that nearly satisfy and fully satisfy the Directed-Acyclic-Graph (DAG) conditions, respectively. Furthermore, upon comparing the obtained relationships and the ground truth, both SCADI and the reproduced CausalVAE successfully capture major causality among the factors. (See Fig.\ref{fig:doset} for SCADI, and Appendix \ref{appendix:exp} for CausalVAE.) Additionally, Table \ref{tab:dag} shows DAGness of $A$ after training, indicating that SCADI, trained to ensure DAG in both the observer and interpreter, exhibits the lowest DAGness. 

Subsequently, we compute the LQ scores for SCADI and nd-SCADI, which are equipped with an observer structure (See Table \ref{tab:ox}.). As shown in Table \ref{tab:main}, SCADI, with additional DAGness imposed on the observer, demonstrated a higher average LQ score in comparison to nd-SCADI. This observation suggests that DAGness aids the observer in anchoring its labels to the underlying factors effectively.

\begin{table}[ht!]
\begin{floatrow}
\floatsetup{captionskip=4pt}
\ttabbox[0.7\Xhsize]
  {\centering
   \tabcolsep=0.1cm
    \renewcommand{\arraystretch}{1.1}
    \small
    \begin{tabular}{ccccccc}
    \hline
    \multicolumn{2}{c}{\multirow{2}{*}{}}          & \multicolumn{4}{c}{\textbf{factor of variation}} & \multirow{2}{*}{\textbf{average LQ}} \\ \cline{3-6}
    \multicolumn{2}{c}{}                           & shad length  & ligth pos  & pendulum  & shad loc &                                      \\ \hline \hline
    \multirow{2}{*}{\textbf{nd-SCADI}} & $u$ index & 3            & 0          & 2         & 3        & \multirow{2}{*}{0.6744}              \\
                                       & LQ        & 0.5915       & 0.4901     & 0.6485    & 0.9675   &                                      \\ \hline
    \multirow{2}{*}{\textbf{SCADI}}    & $u$ index & 0            & 0          & 2         & 3        & \multirow{2}{*}{0.5698}              \\
                                       & LQ        & 0.7401       & 0.6216     & 0.6167    & 0.3007   &                                      \\ \hline
    \end{tabular}}%
    {\caption{Evaluation of unsupervised causal disentanglement methods}\label{tab:main}}%
\hspace{0.3cm}
\ttabbox[\Xhsize]
   {\centering
    \tabcolsep=0.0cm
    \renewcommand{\arraystretch}{1.3}
    \small
    \begin{tabular}{cc}
    \hline
                    & \textbf{DAGness} \\ \hline \hline
    CausalVAE       & 0.5298   \\ \hline
    unsup-CausalVAE & 0.4745   \\ \hline
    nd-SCADI        & 9.7837   \\ \hline
    \textbf{SCADI}           & \textbf{0.1359}   \\ \hline
    \end{tabular}}%
    {\caption{Final DAGness}\label{tab:dag}}%
\end{floatrow}
\end{table}%

\section{Conclusion}
\label{conc}
In conclusion, this paper has endeavored to propose a methodology capable of conducting both observation and interpretation in an unsupervised manner. SCADI is able to capture major causality among factors effectively, and showed better disentanglement result than the other fully unsupervised settings. We hope that our work contributes to future research that aims to achieve unsupervised causal disentanglement.
\bibliographystyle{unsrt}

\section*{Acknowledgement}
This work originated as part of the Electrical \& Electronic Engineering Capstone project at Yonsei University. We are grateful to Jeongryong Lee and Dosik Hwang, whose insightful discussions, support, and encouragement enabled this project. 

\newpage
\bibliography{reference}

\newpage
\begin{appendices}

\section{Implementation details} \label{appendix:imp}

We assign  a 16-dimension latent space ($\alpha=4, c=4$) for the output of the shared encoder. A batch size of 512 and 500 epochs for training are chosen as default. Adam optimizer is adopted with a learning rate of 0.001 for the \textit{observer} and 0.0003 for the \textit{interpreter}. The default DAG constraint for the \textit{observer} is set to $6H(A)+1H(A)^2$, which is twice as large as the DAG constraint used in the \textit{interpreter}, i.e.  $3H(A)+0.5H(A)^2$ is used for \textit{interpreter} which is a default setting of CausalVAE\cite{causalvae}. The $\beta$ parameter for the \textit{observer} is set to 20, while the default setting for the \textit{interpreter} is 4\cite{causalvae}. Every network in the model consists of linear layers with the ELU\cite{elu} activation functions as CausalVAE\cite{causalvae} did. Details of the encoder and decoder architectures are shown in Table \ref{tab:enc} and Table \ref{tab:dec}. We followed Yang et al.\cite{causalvae} for further details.

\begin{figure}[ht!]\TopFloatBoxes
\floatsetup{captionskip=4pt}
\begin{floatrow}
\ttabbox[]
  {\centering
   \tabcolsep=0.05cm
    \renewcommand{\arraystretch}{1.4}
    \small
    \begin{tabular}{cc}
    \hline
    Shared encoder           & Observation encoder               \\ \hline
    Linear$(W*H*C,900)$      & Linear$(\alpha*c,\alpha*c)$ \\
    ELU( )                    & ELU( )                       \\
    Linear$(900,300)$        & Linear$(\alpha*c,\alpha*c)$ \\
    ELU( )                    & ELU( )                       \\
    Linear$(300,2*\alpha*c)$ & Linear$(2*c)$              \\ \hline
    \end{tabular}}
    {\caption{Encoders} \label{tab:enc}}

    \ttabbox[]
    {\centering
    \tabcolsep=0.05cm
    \renewcommand{\arraystretch}{1.4}
    \small
    \begin{tabular}{cc}
    \hline
    Observation decoder & Interpreteration decoder$^a$ \\ \hline
    Linear$(c,300)$           & Linear$(\alpha,300)$       \\
    ELU( )                    & ELU( )                       \\
    Linear$(300,300)$         & Linear$(300,300)$            \\
    ELU( )                    & ELU( )                       \\
    Linear$(300,1024)$        & Linear$(300,1024)$           \\
    ELU( )                    & ELU( )                       \\
    Linear$(1024,1024)$       & Linear$(1024,W*H*C)$         \\
    ELU( )                    &                              \\
    Linear$(1024,W*H*C)$      &                              \\ \hline
    \end{tabular}
    \footnotesize{$^a$ Each factor has seperated interpreter decoder, i.e. Total number of interpreter decoder SCADI has is $c$.}}%
    {\caption{Decoders}
    \label{tab:dec}}%
\end{floatrow}
\end{figure}%

\section{Experiment details} \label{appendix:exp}
\begin{figure}[h!]
\centering{\includegraphics[width=11cm]{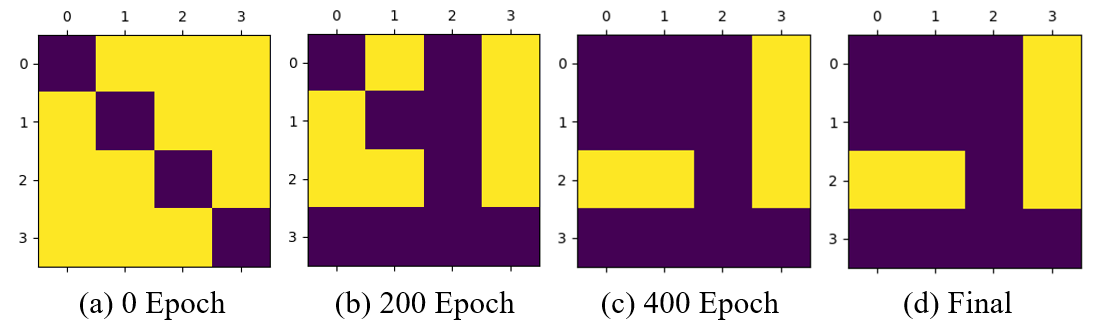}}
\caption{Progress of learning adjacency matrix.}
\label{fig:epochiter}
\end{figure}

While our default training duration comprised 500 epochs, the progress of learning the adjacency matrix $A$ is illustrated in Fig. \ref{fig:epochiter}. In Epoch 0, we initialize the diagonal elements to 0 and all the others to $0.5$ so that initialized matrix $A$ looks like Fig.\ref{fig:epochiter} (a) after rounding up. Fig.\ref{fig:epochiter}(b) still has a bidirectional relationship, but it almost satisfies a DAG. After more iterations, Fig.\ref{fig:epochiter} (c) and (d) satisfy DAG. 

\begin{figure}[htb!]
\centering{\includegraphics[width=0.80\textwidth]{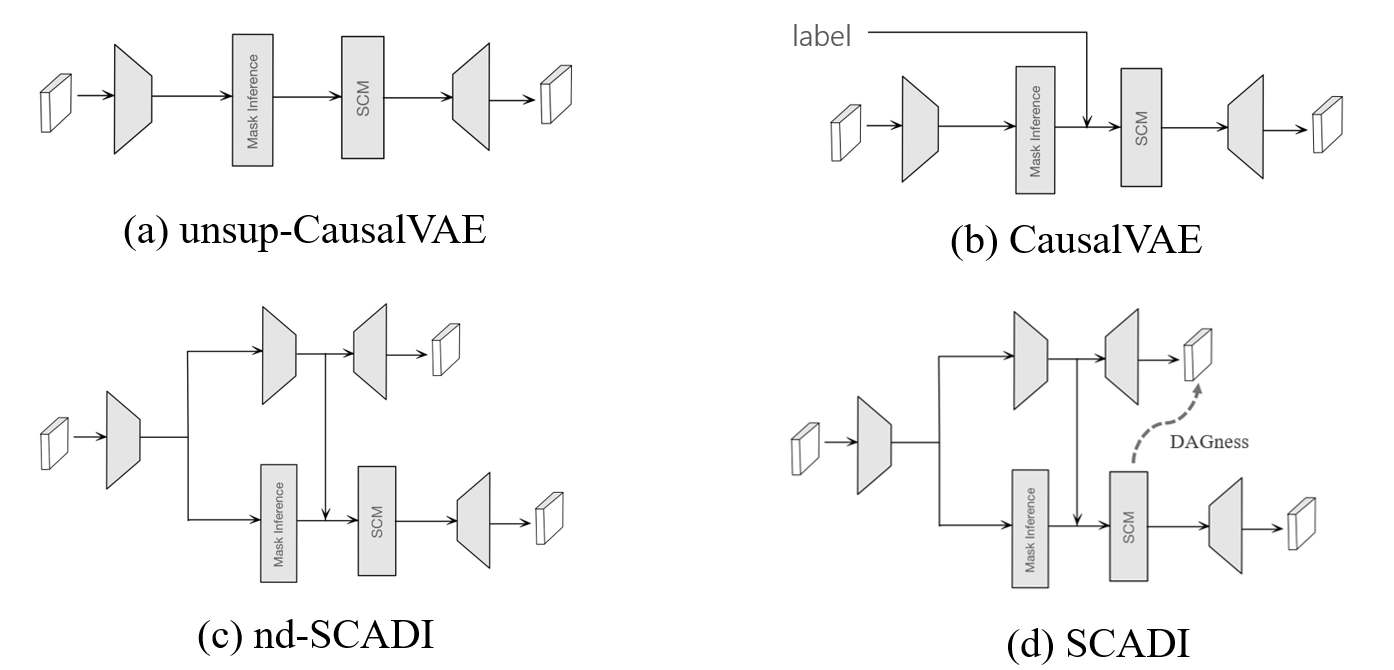}}
\caption{Architectures used in Sec.\ref{sec:unsup}}
\label{fig:models}
\end{figure}

During our experiment in Section\ref{sec:unsup}, four different structures were compared, and Fig.\ref{fig:models} briefly shows the differences among them. 

\begin{figure}[htb!]
\centering{\includegraphics[width=\textwidth]{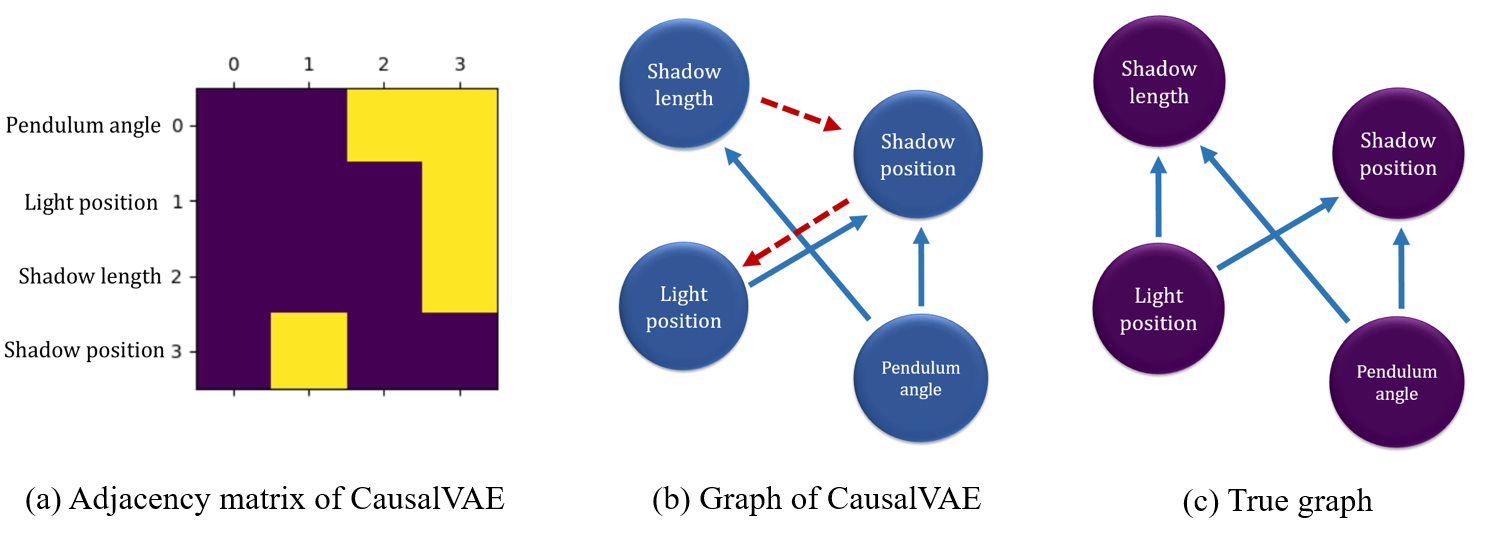}}
\caption{Additional anaylsis on CausalVAE}
\label{fig:causalvae}
\end{figure}

Furthermore, Fig \ref{fig:causalvae} shows additional results from the reproduced CausalVAE\cite{causalvae}. As it operates with weakly supervised labels, there is no need for a label-finding process. The order of factors in the latent variable is predefined: arranged as \textit{pendulum angle, light position, shadow length, and shadow location}. During the reproduction of CausalVAE results, we referenced the default settings proposed by Yang et al.\cite{causalvae}. While not in perfect alignment with the ground truth, it is discernible that the model successfully captured significant causality among the factors.

\section{Ablation study on DAGness}\label{appendix:abl}
\begin{figure}[htb!]
\centering{\includegraphics[width=12cm]{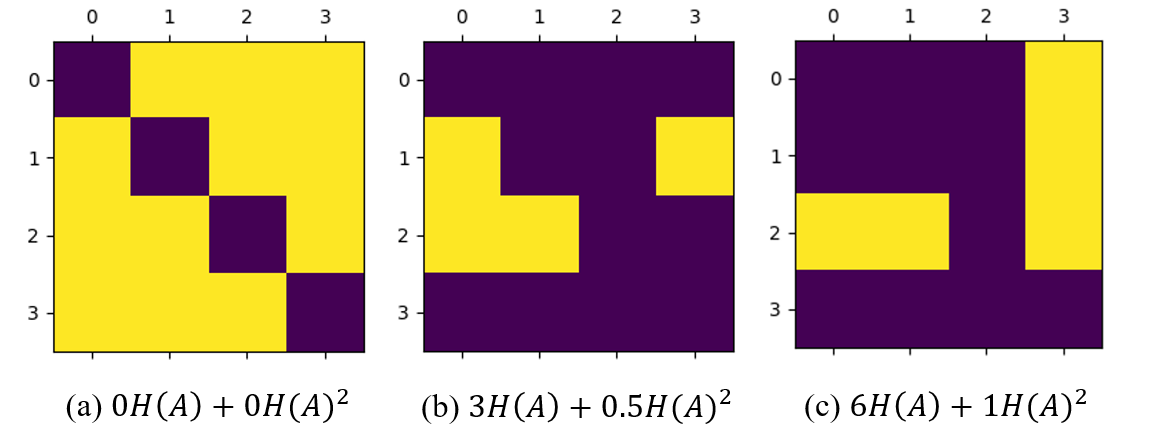}}
\caption{Comparing the adjacency matrices with varying degrees of DAGness}
\label{fig:dagcompare}
\end{figure}

This experiment aims to investigate whether the additional regularization imposed on the \textit{observer}, referred to as DAGness, leads to the generation of higher-quality labels. To investigate the impact of varying levels of DAGness on causal disentangling, we conducted an experiment by giving different conditions: 1) No DAGness to the observer, 2) Half the amount of DAGness compared to our default setting, and 3) SCADI with the default setting. Table \ref{tab:abl1} shows quantitative result and Fig. \ref{fig:dagcompare} shows the adjacency matrices.
\begin{table}[ht!]
\ttabbox[]
  {\centering
   \tabcolsep=0.1cm
    \renewcommand{\arraystretch}{1.2}
    \small
    \begin{tabular}{cccccccc} 
        \hline
        \multirow{2}{*}{\textbf{}}                 & \multirow{2}{*}{} & \multicolumn{4}{c}{\textbf{factor of variation}}                                                                                                                                & \multirow{2}{*}{\textbf{average LQ}} & \multirow{2}{*}{\textbf{DAGness}} \\ \cline{3-6}
                                                   &                   & \multicolumn{1}{c}{shad length} & \multicolumn{1}{c}{light pos} & \multicolumn{1}{c}{pendulum} & shad loc &                                      &                                         \\ \hline \hline
        \multirow{2}{*}{\textbf{$0H(A)+0H(A)^2$}}  & $u$ index    & \multicolumn{1}{c}{3}                      & \multicolumn{1}{c}{0}                       & \multicolumn{1}{c}{2}                       & 3                        & \multirow{2}{*}{0.6744}              & \multirow{2}{*}{9.7387}                 \\
                                                   & LQ       & \multicolumn{1}{c}{0.5915}                 & \multicolumn{1}{c}{0.4901}                  & \multicolumn{1}{c}{0.6485}                  & 0.9675                   &                                      &                                         \\ \hline
        \multirow{2}{*}{\textbf{$3H(A)+0.5H(A)^2$}} & $u$ index    & \multicolumn{1}{c}{0}                      & \multicolumn{1}{c}{0}                       & \multicolumn{1}{c}{0}                       & 3                        & \multirow{2}{*}{0.7581}              & \multirow{2}{*}{0.0942}                 \\
                                                   & LQ       & \multicolumn{1}{c}{1.0366}                 & \multicolumn{1}{c}{0.3373}                  & \multicolumn{1}{c}{1.3862}                  & 0.2755                   &                                      &                                         \\ \hline
        \multirow{2}{*}{\textbf{$6H(A)+H(A)^2$}}   & $u$ index    & \multicolumn{1}{c}{0}                      & \multicolumn{1}{c}{0}                       & \multicolumn{1}{c}{2}                       & 3                        & \multirow{2}{*}{0.5698}              & \multirow{2}{*}{0.1359}                 \\
                                                   & LQ       & \multicolumn{1}{c}{0.7401}                 & \multicolumn{1}{c}{0.6216}                  & \multicolumn{1}{c}{0.6167}                  & 0.3007                   &                                      &                                         \\ \hline 
    \end{tabular}}%
    {\caption{Overall results with varying degree of DAGness}\label{tab:abl1}}%
\end{table}%

In SCADI without imposing DAGness on the \textit{observer}, all factors are causallly entangled, as can be seen in Fig.\ref{fig:dagcompare}, even though labels from label-finding process do not overlap significantly. This indicates that the \textit{observer} could not generate strong labels without DAGness , meaning that the factors are not anchored to the distribution of the true underlying factors. Imposing half the amount of DAGness allows the \textit{observer} to generate a directed acyclic adjacency matrix. However, Table \ref{tab:abl1} shows that the underlying factors are poorly disentangled, showing overlapped labels in the observed latent space. SCADI with a proper amount of DAGness not only has a directed acyclic adjacency matrix but also has the best average LQ score, indicating that both the \textit{observer} and \textit{interpreter} operate well.

\end{appendices}

\end{document}